\documentclass[runningheads]{llncs}

\usepackage{accv}

\usepackage{accvabbrv}

\usepackage{graphicx}
\usepackage{booktabs}

\usepackage[accsupp]{axessibility}  

\usepackage{hyperref}

\usepackage{orcidlink}

\usepackage{pgfplots}
\pgfplotsset{compat=1.18}
\begin{document}

\title{Towards a Joint Khmer Text Recognition and Word Segmentation } 

\titlerunning{Abbreviated paper title}

\author{Marry Kong\inst{1} \and Rina Buoy\inst{1,2}\orcidID{0000-0002-6960-4262} \and Sovisal Chenda\inst{1} \and Nguonly Taing\inst{1} \and Masakazu Iwamura\inst{2} \and Koichi Kise\inst{2}}
%
\authorrunning{Kong et al.}
%
\institute{Techo Startup Center, Ministry of Economy and Finance, Phnom Penh, Cambodia \and Osaka Metropolitan University, Osaka, Japan}

\maketitle

\begin{abstract}
  Text recognition, or extracting electronic text from document images, has been indispensable for knowledge retrieval tasks, such as retrieval-augmented generation (RAG). For Khmer, extracted text is subject to an extra word segmentation step, as Khmer does not use any visible word delimiters to denote word boundaries. Thus, a recognition-then-segmentation pipeline for Khmer requires two separate sequential models; this is not only error-prone but also adds significant latency for large-scale document processing. This paper proposes a novel joint Khmer text recognition and word segmentation framework in a unified model. The proposed model, using a connectionist-temporal-classification (CTC) decoder for fast, parallel decoding, can be instructed to recognize Khmer text with ($b=1$) and without ($b=0$) word segmentation. Experimental results on different benchmark datasets of different document modalities (document, scene, and handwritten images) show that the proposed model can not only recognize characters in document images but also locate word boundaries, removing the need for an extra word segmentation step in a conventional sequential pipeline.  
\end{abstract}

\section{Introduction}
\label{sec:intro}

Central to knowledge retrieval tasks, such as retrieval-augmented generation (RAG)~\cite{lewis2020retrieval}, is text recognition, which is the process of extracting electronic text from document images. For Khmer, an extra word segmentation step is often needed to segment extracted text into words in order to facilitate a hybrid lexical-vector search or apply any post-recognition spell-checking, as Khmer does not use any visible word delimiters in its writing system~\cite{chea2015khmer}.

A common pipeline for Khmer text recognition and word segmentation requires two separate models applied in sequence. Separately, Khmer text recognition and word segmentation are relatively well-explored. Many text recognition methods~\cite{buoy2023toward,valy2020data,nom2024khmerst,nom2025wildkhmerst,buoy2024language,kong2026towards}  have been proposed for the accurate recognition of Khmer characters. Similarly, the task of Khmer word segmentation has also been addressed in multiple previous works~\cite{chea2015khmer,bi2014khmer,kaing2022towards}. However, chaining both models together in a sequential pipeline is not only error-prone but also introduces significant latency when processing a large batch of document images in industrial settings. Furthermore, no previous research has focused on addressing joint text recognition and word segmentation, at least for Khmer text.

In this paper, we propose the first joint \textbf{K}hmer \textbf{t}ext \textbf{r}ecognition and \textbf{w}ord \textbf{s}egmentation (KTRWS) approach that performs both recognition and segmentation of Khmer text in a single model. The proposed approach is fast, as it is based on a connectionist-temporal-classification~\cite{graves2006connectionist} (CTC) decoder for parallel character recognition. In addition, the proposed model can be instructed to recognize Khmer text with ($b=1$) and without ($b=0$) word segmentation. Thus, there is no need for an extra word segmentation step, making the proposed approach particularly suitable for real-world, low-latency, large-scale applications. Since existing Khmer text recognition datasets do not have word boundary annotations, we synthetically generate a new open dataset\footnote{Dataset repository - removed for review} of Khmer text line images with word boundary annotations for model training. 
Our contributions can be summarized as follows: 
\begin{enumerate}
    \item We propose the KTRWS method, which performs joint Khmer text recognition and word segmentation in a single model, along with the first Khmer textline recognition dataset with word boundary annotations.
    \item By performing joint text recognition and word segmentation with a parallel CTC decoder, the proposed method achieves significantly lower latency for this combined task.
    \item Experimental results show that the proposed method can not only recognize Khmer characters accurately but can also eliminate the need for an extra word segmentation step.
\end{enumerate}

\section{Related Work}

\subsubsection{Khmer Text Recognition}

\begin{figure}[tb]
  \centering
  \includegraphics[width=0.6\textwidth]{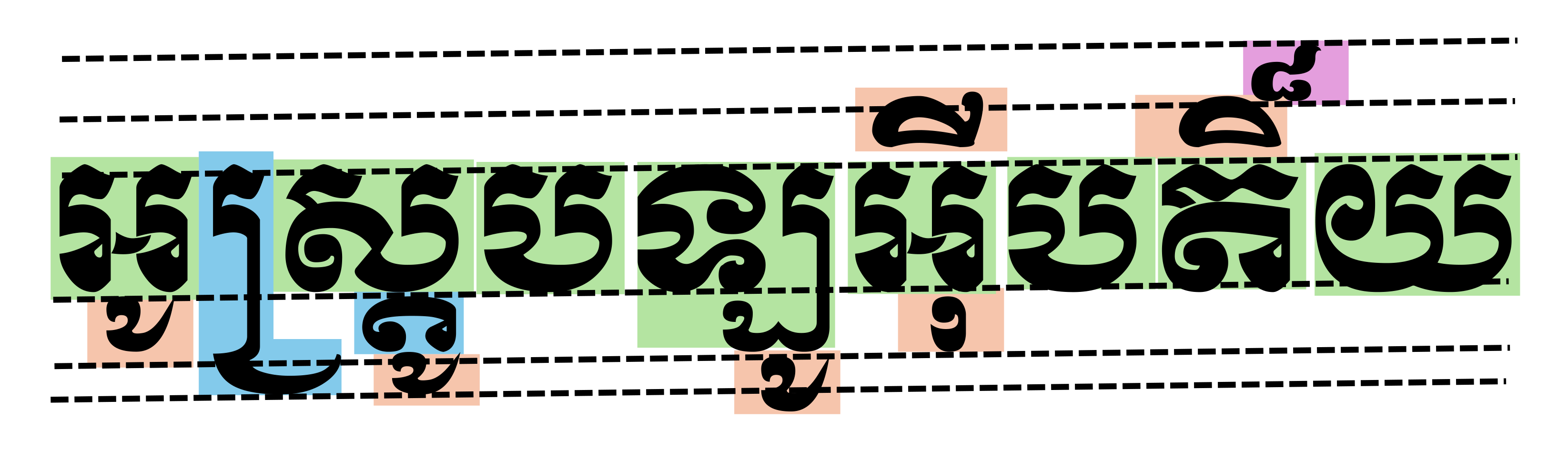}
  \caption{The multi-layered layout of Khmer text, highlighting complex character stacking and ligatures. Green: base consonant. Blue: consonant subscript. Orange: dependent vowel. Purple: diacritic. Best viewed in color.}
  \label{fig:khmerlayout}
\end{figure}

Khmer script, as shown in Figure~\ref{fig:khmerlayout}, is one of the most complex writing systems, characterized by multi-layered character stacking, absence of visible word delimiters, and a large inventory of characters, including 33 consonants, 37 dependent and independent vowels, and eight diacritics~\cite{buoy2023toward,valy2020data}. Despite some recent research progress, Khmer text recognition is a relatively under-explored yet maturing field because of limited public training and benchmark datasets. Early Khmer text recognition methods~\cite{chey2006khmer,kruy2010preliminary,sok2014support} relied on classical feature extractors, such as wavelet descriptors, and classical machine learning methods, such as support vector machines (SVM). Such methods often require explicit character segmentation and are unable to generalize robustly. Deep-learning-based approaches have been proven to achieve superior performance across document modalities (i.e., scene vs. document). Keo et al.~\cite{keo2024state} adopted two prominent Latin-based methods for the task of Khmer scene text recognition on their published dataset using a convolutional feature extractor and two different decoding mechanisms, including a CTC-based decoder and an attention-based decoder. Similarly, Buoy et al.~\cite{buoy2022khmer} addressed the task of Khmer document text recognition using a sequence-to-sequence attentional network. On the other hand, Valy et al.~\cite{valy2020data,valy2017new} proposed multiple deep-learning-based approaches for recognizing Khmer characters and words on historical palm leaf manuscript documents.

Various recent state-of-the-art (SoTA) methods~\cite{buoy2023toward,buoy2025addressing,kong2026towards} for Khmer text recognition have incorporated Transformers~\cite{vaswani2017attention} for both feature extraction and character decoding, and proposed a Khmer-native tokenization scheme called Khmer character clusters (KCC) instead of conventional character-level tokenization. As a result, character error rates (CERs) across public benchmark datasets, including scene, document, and handwritten images, have been significantly reduced.

\subsubsection{Khmer Word Segmentation}
Since Khmer does not use any visible delimiters to denote word boundaries, word segmentation is a necessary prior step for any downstream natural language processing (NLP) tasks, such as part-of-speech (POS) tagging, named entity recognition (NER), spell checking, lexical search, and so on. Nonetheless, there is no standard definition of a word in Khmer, nor a standard corpus for this task~\cite{kaing2022towards}. As a result, researchers often construct their own corpora and train their own models.

The early work~\cite{bi2014khmer} on Khmer word segmentation was based on a dictionary-based bidirectional maximal matching (BiMM) approach. Despite being simple (requiring no training) and lightweight, the BiMM approach is unable to handle out-of-vocabulary (OOV) words. To address this limitation, Chea et al.~\cite{chea2015khmer} proposed a statistical machine learning approach using a conditional random field (CRF) algorithm. The authors manually constructed training and evaluation datasets using word-level and compound-level (i.e., prefixes and suffixes) annotations and trained the CRF-based model. The CRF-based model achieved an accuracy of ~98.5\% and has become a de facto baseline model.

In addition, Buoy et al.~\cite{buoy2020khmer} proposed a neural Khmer word segmentation network using a bidirectional long short-term memory (BiLSTM) architecture to improve contextual information modeling. The authors proposed two variants using character-level and character-cluster-level tokenization schemes. Recognizing that word segmentation and POS tagging can be jointly trained, Kaing et al.~\cite{kaing2022towards} proposed a joint word segmentation and POS tagging method.

In summary, while both word segmentation and text recognition for Khmer have significantly evolved along their respective, seemingly different trajectories, both tasks meet and complement each other in downstream applications, such as knowledge retrieval, where electronic texts are recognized from document images and words are segmented for ingestion and retrieval. Thus, unifying both tasks within a single model can lead to significant latency reduction, which is of great importance for practical applications. To this end, this paper proposes a novel KTRWS approach.

\section{The Proposed Method}

\begin{figure}[tb]
  \centering
  \includegraphics[width=0.8\textwidth]{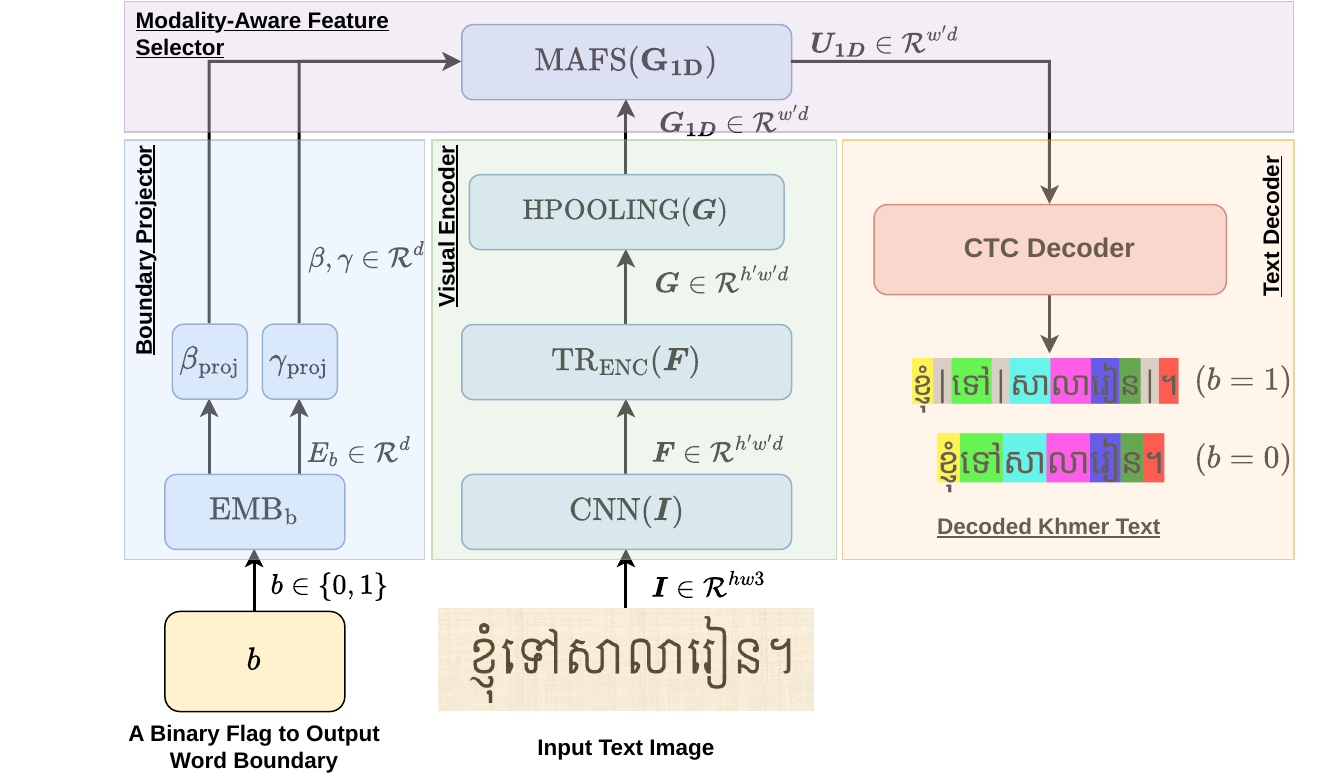}
  \caption{The proposed KTRWS framework. Best viewed in color.}
  \label{fig:method}
\end{figure}

As shown in Figure~\ref{fig:method}, our proposed KTRWS framework comprises four key components: a word boundary projector, a visual encoder, a modality-aware feature selector (MAFS), and a text decoder. The boundary projector embeds and projects a binary flag ($b$) indicating whether to output word boundaries. The visual encoder extracts both visual and sequential features from an input image. The MAFS fuses word boundary projections and visual-temporal features, and adapts according to a specific modality (i.e., scene, handwritten, and document text). Finally, the text decoder uses a CTC decoder to output Khmer tokens with ($b=1$) or without word boundary tokens ($b=0$).

\subsubsection{Word Boundary Projector Module}

This module embeds and projects a binary flag indicating whether to output word boundaries, and returns two projection vectors, $\boldsymbol{\beta} \in \mathbb{R}^d$ and $\boldsymbol{\gamma} \in \mathbb{R}^d$. These projection vectors are fused with the adapted features from the encoder before Khmer character decoding. With this design, the model can be instructed to output word boundaries or not during inference. Mathematically, the word boundary projector module can be expressed as

\noindent
\begin{align}\label{eq:1}
\boldsymbol{E_b} &=\operatorname{\mathrm{LINEARLAYER}}(b)\\
\boldsymbol{\beta} &=\operatorname{\mathrm{LINEARLAYER}}(\boldsymbol{E_b})\\
\boldsymbol{\gamma} &=\operatorname{\mathrm{LINEARLAYER}}(\boldsymbol{E_b}),
\end{align}
where $b \in \{0,1\}$ is a binary variable indicating whether to output word boundaries, $\operatorname{LINEARLAYER}$ is a dense neural network layer, $\boldsymbol{E}_b \in \mathbb{R}^d$ is an embedding vector of $b$, and $\boldsymbol{\beta}$ and $\boldsymbol{\gamma}$ are the resulting projection vectors.

\subsubsection{Visual Encoder}\label{visual_encoder} 

The visual encoder is designed to extract the visual-temporal features required for accurate character recognition. For a given RGB image ($\boldsymbol{I}$), our visual encoder consists of a base convolutional network for two-dimensional (2D) visual features ($\boldsymbol{F}  \in \mathbb{R}^{w'\times h' \times d}$) and a Transformer-based encoder network for visual-temporal features ($\boldsymbol{G}  \in \mathbb{R}^{w'\times h' \times d}$). Since the CTC decoder expects a sequence of one-dimensional (1D) features ($\boldsymbol{G}_\text{1D} \in \mathbb{R}^{w'\times d}$ ) along the width direction, the extracted visual-temporal features are averaged over the height direction. Mathematically, our visual encoder can be expressed as

\noindent
\begin{align}\label{eq:2}
\boldsymbol{F} &=\operatorname{\mathrm{CNN}}(\boldsymbol{I})\\
\boldsymbol{G} &=\operatorname{\mathrm{TR}_{ENC}}(\boldsymbol{F})\\
\boldsymbol{G_{1D}} &=\operatorname{\mathrm{HPOOLING}}(\boldsymbol{G}),
\end{align}
where $\operatorname{CNN}$, $\operatorname{TR}_\text{ENC}$, and $\operatorname{HPOOL}$ are the base convolutional network, the Transformer encoder network, and the height-averaging operator, respectively; $\boldsymbol{I}$ is an input RGB image; and $d$ (= 512) is the feature dimension.

\subsubsection{Modality-Aware Feature Selector}\label{visual_encoder} 

To enable the proposed framework to robustly recognize Khmer text across different modalities (e.g., scene, handwritten, and document), we modify the modality-aware feature selector~\cite{kong2026towards} (MAFS) to incorporate the word boundary projections (i.e., $\boldsymbol{\beta}$ and $\boldsymbol{\gamma}$). Specifically, this module computes an average vector $\boldsymbol{z} \in \mathbb{R}^d$ using the $\operatorname{GPOOL}$ operator, given $\boldsymbol{G}_\text{1D}$. The module then uses $\operatorname{ROUTER}$ to compute a distribution over $n$ modality sources ($\boldsymbol{r} \in \mathbb{R}^n$), from which modality-adapted features ($\boldsymbol{H}_{\text{1D}} \in \mathbb{R}^{d\times n}$) from $\operatorname{ADAPTER}$ are marginalized to yield the final features ($\boldsymbol{U}_\text{1D} \in \mathbb{R}^{w'\times d}$), which are then subject to feature-wise linear modulation~\cite{perez2018film} (FiLM), scaled and shifted by $\boldsymbol{\gamma}$ and $\boldsymbol{\beta}$, for output token generation. Mathematically, the modified MAFS can be expressed as

\noindent
\begin{align}\label{eq:3}
\boldsymbol{z} &=\operatorname{\mathrm{GPOOLING}}(\boldsymbol{G_{1D}})\\
\boldsymbol{r} &=\operatorname{\mathrm{ROUTER}}(\boldsymbol{z})\\
\boldsymbol{H_{1D,k}} &=\operatorname{\mathrm{ADAPTER_k}}(\boldsymbol{G_{1D}})\\
\boldsymbol{U_{1D}} &=\boldsymbol{\gamma }\odot (\boldsymbol{H_{1D}}\cdot \boldsymbol{r})+\boldsymbol{\beta},
\end{align}
where $\boldsymbol{U}_\text{1D}$ is the input feature map for the CTC decoder to generate output Khmer tokens. We adopt the same tokenization scheme as~\cite{kong2026towards,buoy2025addressing}, which operates at the character-cluster level rather than the character level.

\subsubsection{Text Decoder}\label{text_decoder} 
As shown in Figure~\ref{fig:method}, we utilize a CTC decoder for parallel Khmer token generation. Given the input feature map $\boldsymbol{U}_\text{1D} \in \mathbb{R}^{w' \times d}$ from the MAFS module, the decoder projects each of the $w'$ frames onto the augmented vocabulary $\mathcal{V}' = \mathcal{V} \cup \{\varnothing\} \cup \{\texttt{\texttt{u200b}}\}$, where $\mathcal{V}$ is the set of character-cluster tokens, $\varnothing$ is the blank symbol, and $\texttt{\texttt{u200b}}$ is a word boundary marker token (invisible when rendered). A softmax classifier produces a per-frame distribution as

\noindent
\begin{align}\label{eq:4}
p_i = \operatorname{\mathrm{SOFTMAX}}(\boldsymbol{U_{1D,i}}),
\end{align}
where $p_i(c)$ denotes the probability of emitting token $c$ at frame $i$.

For a given target sequence $\boldsymbol{y}$,  a $\mathcal{L}_{\text{CTC}}$ loss is given by

\noindent
\begin{align}\label{eq:5}
\mathcal{L}_{\text{CTC}} = -\!\!\sum_{(I,\,y)\in\mathcal{D}} \log p(\boldsymbol{y} \mid \boldsymbol{U_{1D}}), \tag{14}
\end{align}
where $\log p(\boldsymbol{y} \mid \boldsymbol{U_1D})$ is obtained by marginalizing over all alignments that collapse to $\boldsymbol{y}$. At test time, we use greedy (best-path) decoding, which approximates the most probable sequence by selecting the most likely token independently at each frame and collapsing repeated tokens and $\varnothing$.

\noindent
\begin{align}\label{eq:6}
\hat{\pi}_i = \arg\max_{c \in \mathcal{V}'} p_i(c), \qquad i = 1, \dots, w'
\end{align}

\subsubsection{Alternative Design for Word Boundary Projector Module}

Alternatively, the boundary embedding vector can be fused via gating. Mathematically, the word boundary projector with gating can be expressed as

\noindent
\begin{align}\label{eq:10}
\boldsymbol{\alpha} &=\sigma(\operatorname{\mathrm{LINEARLAYER}}(\boldsymbol{E_b}))\\
\boldsymbol{p} &=\operatorname{\mathrm{LINEARLAYER}}(\boldsymbol{E_b})\\
\boldsymbol{U_{1D}} &=\boldsymbol{\alpha}\odot (\boldsymbol{H_{1D}}\cdot \boldsymbol{r}) + (1 - \boldsymbol{\alpha})\odot \boldsymbol{p},
\end{align}
where $b \in \{0,1\}$ is a binary variable indicating whether to output word boundaries, $\operatorname{LINEARLAYER}$ is a dense neural network layer, $\boldsymbol{E}_b \in \mathbb{R}^d$ is an embedding vector of $b$, and $\boldsymbol{\sigma}$ is a sigmoid gating function applied element-wise to the embedding vector. 

\section{Datasets \& Experimental Setup}
\label{sec:datasets}

 \begin{figure}[htbp]
    \centering
    \includegraphics[width=0.48\textwidth]{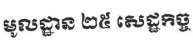}\hfill
    \includegraphics[width=0.48\textwidth]{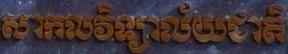}

    \vspace{0.5em}
    \includegraphics[width=0.48\textwidth]{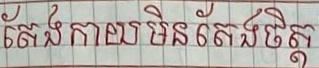}\hfill
    \includegraphics[width=0.48\textwidth]{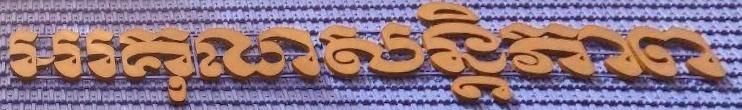}

    \vspace{0.5em}
    \includegraphics[width=0.48\textwidth]{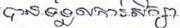}\hfill
    \includegraphics[width=0.48\textwidth]{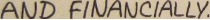}

    \caption{Sample preview images from the existing datasets.}
    \label{fig:existing_images}
\end{figure}

\subsubsection{Datasets} Prior Khmer text recognition methods have relied largely on synthetic training data, which is convenient to generate at scale, whereas real data for the scene and handwritten modalities remains scarce. We adopt the publicly available datasets listed below, comprising real and synthetic Khmer document, scene, and handwritten text images. A summary of all datasets and some preview images are given in Table~\ref{tab:datasets} and Figure~\ref{fig:existing_images}, respectively. 


\begin{description}
  \item[Buoy et al.:~\cite{buoy2023toward}] This dataset comprises about 2.8M synthetic textlines (1.5M document, 1.3M scene)
  rendered with 11 Khmer fonts. This dataset features document and scene modalities.
 
  \item[KHOB~\cite{ekycsolutions2022khmerocr}:] This dataset comprises 1{,}318 textlines manually cropped from scanned low-resolution
  Khmer PDF documents. This dataset is used for document-modality evaluation.
 
  \item[SynthText~\cite{seanghay_synthkhmer10k}:] This dataset comprises 70{,}000 textlines extracted from 10{,}000 synthetic Khmer
  identity card images. This dataset features document modality.
 
  \item[HierText~\cite{long2023icdar}:] This dataset comprises a large-scale Latin printed/handwritten dataset yielding about
  518{,}726 textline images after cropping and removing vertical texts. This dataset is used to learn Latin visual representations (scene and handwritten modalities).
 
  \item[KhmerST~\cite{nom2024khmerst}:] This dataset comprises 3{,}022 real Khmer scene text images (mostly logos and
  billboards) with diverse artistic fonts and challenging imaging conditions;
  This dataset is used for scene-modality evaluation.
 
  \item[WildKhmerST~\cite{nom2025wildkhmerst}:] This dataset comprises 29{,}601 textlines from 10{,}000 real images across Cambodia,
  including artistic, blurred, low-light, curved, and occluded text. This dataset is used for scene
  modality.
 
  \item[KH~\cite{soyvitou_khmerhandwritten4_2k}:] This dataset comprises about 4.2k Khmer and Latin textlines (3{,}991 train,
  211 eval), both handwritten and printed. This dataset features document and handwritten modalities.
 
  \item[GKST~\cite{kong2026towards}:] This dataset comprises 4{,}221 smartphone-captured Khmer scene text images
  (4{,}009 train, 212 eval), annotated from general scenes rather than focused
  close-ups. This dataset features scene modality.
 
  \item[KHT~\cite{kong2026towards}:] This dataset comprises 14{,}168 Khmer handwritten text images (13{,}457 train,
  711 eval) from sources such as birth certificates, exam papers, and notes. This dataset features handwritten modality.
\end{description}

\begin{figure}[htbp]
    \centering
    \begin{subfigure}{\textwidth}
    \centering
        \includegraphics[width=0.8\linewidth]{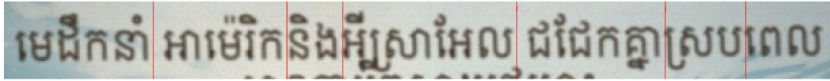}
        \caption{English translation: \emph{The leaders of the United States and Israel discussed while}.}
        \label{fig:img1}
    \end{subfigure}

    \vspace{0.5em}
   
    \begin{subfigure}{\textwidth}
    \centering
        \includegraphics[width=0.8\linewidth]{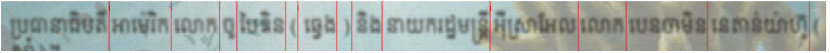}
        \caption{English translation: \emph{The US President Joe Biden (left) and the Israeli Prime Minister Benjamin Netanyahu (}.}
        \label{fig:img2}
    \end{subfigure}

    \vspace{0.5em}
    
    \begin{subfigure}{\textwidth}
    \centering
        \includegraphics[width=0.3\linewidth]{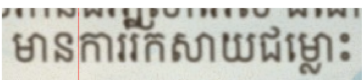}
        \caption{English translation: \emph{The spread of conflict}.}
        \label{fig:img3}
    \end{subfigure}

    \caption{Sample generated images with word boundary annotations (red bar) from our new dataset.}
    \label{fig:newdataset_image}
\end{figure}

Since existing datasets do not provide word boundary annotations, we synthetically generate 60,000 text line images with word boundary annotations. To achieve this, we begin by training a Khmer Transformer-based model for the word segmentation task using the manually labelled dataset provided by Chea et al.~\cite{chea2015khmer}. The textlines, sourced mainly from news articles and segmented by an invisible word boundary marker (i.e., \texttt{\texttt{u200b}}), are rendered synthetically as images for model training. Consequently, the rendered text images are not aesthetically affected by the extra word boundary information. Some preview images from this new dataset with word boundary annotations are provided in Figures~\ref{fig:newdataset_image}.
\begin{table}[t]
  \centering
  \caption{Summary of datasets. Size: train\,/\,eval (--~if none). Purpose:
  Tr~=~training (D), Ad~=~adapting (S\&H), Ev~=~evaluation. Modality:
  D~=~document, S~=~scene, H~=~handwritten.}
  \label{tab:datasets}
  \setlength{\tabcolsep}{6pt}
  \begin{tabular}{lrll}
    \toprule
    Dataset & Size & Purpose & Modality \\
    \midrule
    Buoy et al.   & 2.8M\,/\,--        & Tr     & D, S \\
    KHOB          & --\,/\,1{,}318     & Ad, Ev & D    \\
    SynthText     & 70k\,/\,--         & Tr     & D    \\
    HierText      & 518{,}726\,/\,--   & Tr     & S, H \\
    KhmerST       & --\,/\,3{,}022     & Ad, Ev & S    \\
    WildKhmerST   & 29{,}601\,/\,--    & Ad     & S    \\
    KH            & 3{,}991\,/\,211    & Ad, Ev & D, H \\
    GKST   & 4{,}009\,/\,212    & Ad, Ev & S    \\
    KHT    & 13{,}457\,/\,711   & Ad, Ev & H    \\
    \bottomrule
  \end{tabular}
\end{table}

\subsubsection{Experimental Setup \& Training Strategy}
\label{subsec:experimental-setup}

The base $\operatorname{CNN}$ feature extractor is composed of six sequential ResNet blocks with channel dimensions progressively increasing from 32 to 512, where each block comprises 1×1 and/or 3×3 convolutions repeated multiple times. Spatial downsampling by a factor of $(2,2)$ is performed at ResNet Blocks 1 and 4, while the remaining blocks maintain the spatial resolution. The $\operatorname{TR}_\text{ENC}$ network is configured with an embedding dimension ($d$) of 512, three stacked encoder layers, eight attention heads, a dropout rate of 0.1, and a feed-forward network dimension of 2048 ($4\times d$). The model size and computational complexity of the visual encoder are reported in Table~\ref{encoder_params_flops}. Although there are three discrete modalities, document, scene, and handwritten, real images rarely belong to only one modality. Instead, they can be represented as points within a simplex spanning these three modalities. For example, a text image may be simultaneously both handwritten and scene text. By default, we set this hyperparameter ($n$) to five, providing additional capacity to model hybrid or transitional cases~\cite{kong2026towards}.

Following~\cite{kong2026towards}, we group these datasets by primary modality: \emph{Document (D)} for large-scale
printed data (Buoy et al., SynthText, HierText); \emph{Scene \& Handwritten
(S\&H)} for real data (WildKhmerST, the KH training set, GKST, and KHT); and an
\emph{Evaluation} group comprising KhmerST (S), KHOB (D), the KH evaluation set
(D\&H), GKST (S) and KHT (H). 

\begin{table}[t]
\centering
\caption{The visual encoder's complexity and model size. Params: model parameters. Flops: floating operations on an input of $32\times116$ pixels. }\label{encoder_params_flops}
\begin{tabular}{lrr}
\hline
\textbf{Name} &  \textbf{Params.} &\textbf{Flops} \\
\hline
ResNet backbone (visual features) & 13.0M & 3.2G \\
Transformers encoder (sequential features)& 9.5M & 2.2G  \\
MAFS & 0.80M & 0.15G \\
\hline

\hline
\end{tabular}
\end{table}

The KTRWS models were trained in two phases: a general training phase followed by
a modality-adapting phase. In the \emph{general training phase}, the models were first trained on the
large-scale document (D group) datasets to learn robust visual representations of
Khmer and Latin characters. A cyclic learning rate schedule was employed, with a
minimum value of $10^{-5}$ and a maximum of $10^{-4}$. This phase lasted five full
epochs with a batch size of 32 images. In the \emph{modality-adapting phase}, the trained models then underwent
adaptation on the real scene and handwritten (S\&H group) datasets together with our newly annotated word-boundary dataset for 50 epochs,
using a lower cyclic learning rate schedule with a minimum of $10^{-6}$ and a
maximum of $10^{-5}$. In both phases, the Adam optimizer was used with a gradient
clipping value of 50. To prevent the models from underperforming on printed
document text during this phase, we sampled an equal number of document images and
mixed them with the S\&H images. As a result, the same model can both recognize Khmer text across the document,
scene, and handwritten modalities and, when conditioned on $b = 1$, additionally
output word boundaries.

\section{Results \& Discussion}

\begin{table*}[t]
\centering
\caption{Character error rates (CER) on the evaluation datasets of different modalities (D,S,\&H). * : model-specific tokenizer. Char.: character. KCC: Khmer character cluster. Seg.: joint character recognition and word segmentation. \textbf{Bold}: best. \emph{Italic}: second best. }
\label{tab:reg_accuracy}
\begin{tabular}{lcccccccc}
\toprule
 & Dec. & Tok.  & Seg. & KHOB(D) & KhmerST(S) & KH(D,H) & GKST(S) & KHT(H) \\
\midrule

Surya-OCR~\cite{paruchuri2025surya} & Tr. & * & No & 17.69 & 43.21 & -- & -- & -- \\
Buoy et al.~\cite{buoy2023toward} & Tr. & Char. & No & 3.03 & -- & -- & -- & -- \\
Nom et al.~\cite{nom2024khmerst} & Tr. & *& No  & -- & 17.00 & -- & -- & -- \\
Soy et al.~\cite{soyvitou_khmerhandwritten4_2k} & Tr. & *& No  & -- & -- & 17.00 & -- & -- \\
Buoy et al.~\cite{buoy2025addressing} & Tr. & KCC& No & 2.13 & 7.01 & -- & -- & -- \\
\midrule
Tesseract-OCR~\cite{tesseract_ocr} & CTC & Char. & No & 9.19 & 40.96 & -- & -- & -- \\
Buoy et al.~\cite{buoy2024language} & CTC & KCC & No & 2.33 & -- & -- & -- & -- \\
UKTR~\cite{kong2026towards} & CTC & KCC & No &  2.46 & \emph{3.02} & \emph{5.89} & \textbf{4.41} & \emph{9.52}\\
\midrule
\textbf{KTRWS} ($b=0$) & CTC & KCC & No &  \textbf{1.75} & \textbf{2.77} & \textbf{5.78} & \emph{4.51} & \textbf{8.91}\\
\textbf{KTRWS} ($b=1$) & CTC & KCC & Yes &  \emph{1.79} & 3.04 & 6.06 & 5.01 & 9.64\\


\bottomrule
\end{tabular}
\end{table*}
\subsubsection{Recognition Performance of the KTRWS Model} We begin by comparing the recognition performance of the proposed method in terms of character error rate (CER) across the evaluation datasets of different modalities. Since the evaluation datasets do not have word boundary annotations, CERs are computed by excluding the $\texttt{u200b}$ word boundary marker token.

Table~\ref{tab:reg_accuracy} provides the recognition accuracy comparisons of the proposed models, with and without word boundary outputs, against the SoTA methods. Compared with existing methods, the proposed models either outperform on the KHOB dataset or are competitive on the other datasets (KhmerST, KH, GKST, and KHT). Compared with the recent SoTA UKTR model using the same CTC decoder, our $\textbf{KTRWS}$ ($b=0$) model without outputting word boundaries achieves improved CERs on all datasets except GKST. Specifically, our $\textbf{KTRWS}$ ($b=0$) model (i.e., no word boundary outputs) obtained CERs of 1.75\%, 2.77\%, 5.78\%, 4.51\%, and 8.91\% on KHOB, KhmerST, KH, GKST, and KHT, respectively, versus 2.46\%, 3.02\%, 5.89\%, 4.41\%, and 9.52\% by the CTC-based UKTR model. When outputting word boundaries, our $\textbf{KTRWS}$ ($b=1$) model remains competitive. Specifically, our $\textbf{KTRWS}$ ($b=1$) model obtained CERs of 1.79\%, 3.04\%, 6.06\%, 5.01\%, and 9.64\%. The marginal increases in CER with $b=1$ are expected, as the model needs to generate both normal Khmer tokens and word boundary tokens, which increases the probability of making errors. On the other hand, the significant CER improvements on the KHOB dataset can be attributed to the fact that the new dataset introduced in this study is more aligned with the nature of the KHOB dataset (i.e., long document textlines).

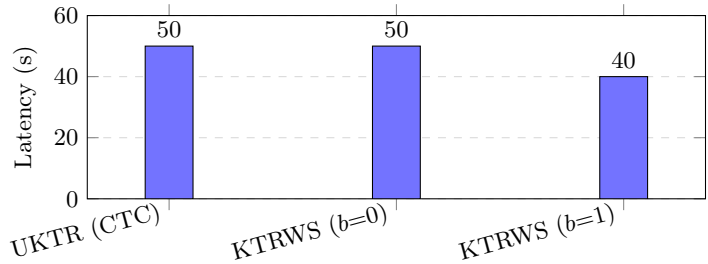
\begin{figure}[t]
  \centering
  \begin{tikzpicture}
    \begin{axis}[
        ybar,
        width=0.8\linewidth,
        height=4cm,
        bar width=18pt,
        ymin=0, ymax=60,
        ylabel={Latency (s)},
        symbolic x coords={ UKTR (CTC), KTRWS ($b{=}0$), KTRWS ($b{=}1$)},
        xtick=data,
        x tick label style={font=\small, rotate=15, anchor=east},
        nodes near coords,
        nodes near coords align={vertical},
        every node near coord/.append style={font=\small},
        enlarge x limits=0.18,
        ymajorgrids=true,
        grid style={dashed, gray!30},
      ]
      \addplot[fill=blue!55] coordinates {
        (UKTR (CTC), 50)
        (KTRWS ($b{=}0$), 50)
        (KTRWS ($b{=}1$), 40)
      };
    \end{axis}
  \end{tikzpicture}
  \caption{Latency (seconds) comparison on for the join recognition and segmentation task on the KHOB evaluation set.}
  \label{fig:latency}
\end{figure}

As shown in Figure~\ref{fig:latency}, our $\textbf{KTRWS}$ ($b=1$) model achieves significantly lower combined recognition and segmentation latency on the KHOB evaluation set than the sequential recognition-then-segmentation approaches (i.e., UKTR or $\textbf{KTRWS}$ ($b=0$)). This is attributed to its parallel, joint decoding nature, which eliminates the need to perform word segmentation as a separate step using an external model such as the Transformer-based Khmer word segmentation model described in Section~\ref{sec:datasets}. It should be noted that the degree of latency reduction is directly proportional to the latency of the word segmentation model used in the sequential pipeline.


\subsubsection{Segmentation Accuracy of the KTRWS Models}

\begin{table*}[t]
\centering
\caption{F1 scores on the evaluation datasets of different modalities (D,S,\&H). * : model-specific tokenizer. Char.: character. KCC: Khmer character cluster. \textbf{Bold}: best. \emph{Italic}: second best.}
\label{tab:reg_accuracy_word}
\begin{tabular}{lcccccccc}
\toprule
 & KHOB(D) & KhmerST(S) & KH(D,H) & GKST(S) & KHT(H) \\
\midrule

\textbf{KTRWS} ($b=0$) &  98.49 & 97.90 & 95.84 & 96.41  & 93.40\\
\textbf{KTRWS} ($b=1$)  &  97.68 &  95.08 & 91.09 & 93.35  & 89.24\\


\bottomrule
\end{tabular}
\end{table*}

\begin{figure}[htbp]
    \centering
    \begin{subfigure}{\textwidth}
    \centering
        \includegraphics[width=\linewidth]{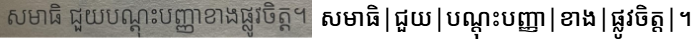}
        \caption{English translation: \emph{Meditation helps to train the mind}.}
        \label{fig:img1_qualitative}
    \end{subfigure}
    \vspace{0.5em}
    \begin{subfigure}{\textwidth}
    \centering
        \includegraphics[width=\linewidth]{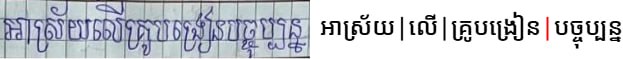}
        \caption{English translation: \emph{Depending the present teachers}.}
        \label{fig:img2_qualitative}
    \end{subfigure}
    \vspace{0.5em}
    \begin{subfigure}{\textwidth}
    \centering
        \includegraphics[width=\linewidth]{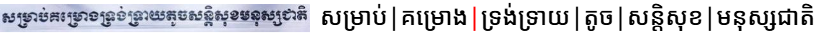}
        \caption{English translation: \emph{For small projects of humanity security}.}
        \label{fig:img3_qualitative}
    \end{subfigure}
    \vspace{0.5em}
    \begin{subfigure}{\textwidth}
    \centering
        \includegraphics[width=\linewidth]{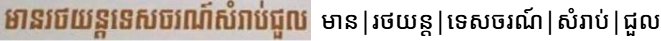}
        \caption{English translation: \emph{Tour vehicles for renting}.}
        \label{fig:img4_qualitative}
    \end{subfigure}
    \vspace{0.5em}
    \begin{subfigure}{\textwidth}
    \centering
        \includegraphics[width=\linewidth]{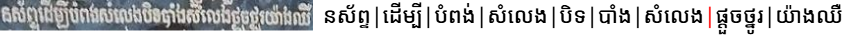}
        \caption{English translation: \emph{For the vocal cords to muffle the painful groans}.}
        \label{fig:img5_qualitative}
    \end{subfigure}
    \vspace{0.5em}
    \begin{subfigure}{\textwidth}
    \centering
        \includegraphics[width=\linewidth]{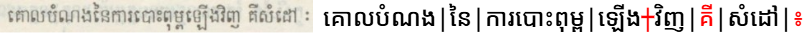}
        \caption{English translation: \emph{For the vocal cords to muffle the painful groans}.}
        \label{fig:img6_qualitative}
    \end{subfigure}
    \caption{Qualitative assessment of recognition and segmentation accuracy on some selected cases. Red bar: missing word boundary. Strided red bar: extra word boundary. Red: recognition mistake.}
    \label{fig:qualitative_evaluation}
\end{figure}

We evaluate word segmentation using F1 on the evaluation datasets.
For each textline image, let $\mathcal{R}$ and $\mathcal{H}$ denote the reference
and hypothesis word sequences, respectively. There are two possible cases:

\begin{itemize}
    \item Without OCR errors: when the concatenated character sequences agree ($\bigcup\mathcal{R} =
        \bigcup\mathcal{H}$), we compare the sets of cumulative character-offset
        boundaries $B_{\mathcal{R}}$ and $B_{\mathcal{H}}$ directly and compute true positive ($\mathrm{TP}$), false positive ($\mathrm{FP}$), and false negative ($\mathrm{FN}$).
        
    \item With OCR errors: When OCR errors cause a character mismatch, boundary offsets are no longer
comparable. We fall back to word-level multiset overlap $\mathrm{TP} = \sum_{w}{\min(c_{\mathcal{R}}(w),\,c_{\mathcal{H}}(w))}$,
where $c(\cdot)$ denotes word counts, with FP and FN defined analogously.
\end{itemize}

TP, FP, and FN are summed across all textline images and used to compute
micro-averaged F1 score.

Since none of the evaluation datasets originally have word boundary annotations, we used our trained word segmentor discussed in Section~\ref{sec:datasets} to segment the ground-truth texts. The same applies to the model predictions of our $\textbf{KTRWS}$ ($b=0$) model (i.e., without word boundaries). In contrast, the model predictions of our $\textbf{KTRWS}$ ($b=1$) model already include word boundaries. Table~\ref{tab:reg_accuracy_word} shows that the segmentation accuracy of our $\textbf{KTRWS}$ ($b=1$) model degrades as the document modality transitions from document to scene and handwritten text. This is because the document dataset (KHOB) mainly comprises long text-line images with complete sentences, which are not hard to recognize and provide sufficient contextual cues for word segmentation. On the other hand, the scene and handwritten cases (KhmerST, KH, GKST, and KHT) mainly comprise single words, short phrases, or non-textual content (e.g., numbers), which are out-of-context and challenging for both recognition and word boundary identification.

\subsubsection{Qualitative Assessment}
Figure~\ref{fig:qualitative_evaluation} provides a qualitative assessment of the recognition and segmentation outputs from our $\textbf{KTRWS}$ ($b=1$) model on selected cases. Consistent with the above observations, while recognition errors are infrequent, except in Figure~\ref{fig:img6_qualitative} (i.e., a historically degraded image), segmentation errors (mainly missing word boundaries) occur more frequently on scene and handwritten text images, as in the cases of Figures~\ref{fig:img2_qualitative},~\ref{fig:img3_qualitative},~\ref{fig:img5_qualitative}, and~\ref{fig:img6_qualitative}.

\subsubsection{Visual Grounding \& Alternative Design for Word Boundary Projector Module } 

Another benefit of the proposed joint recognition and word segmentation method is the visual grounding of recognized words. Thanks to the CTC decoder predicting $p_i(c)$ for each feature frame along the width direction, it is possible to locate the word boundary tokens and map them to their actual positions on the input image. As a result, each predicted word can be accurately visually grounded.

\begin{figure}[htbp]
    \centering
    \begin{subfigure}{\textwidth}
    \centering
        \includegraphics[width=0.5\linewidth]{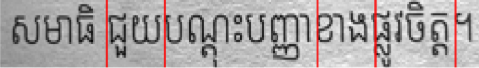}
        \caption{English translation: \emph{Meditation helps to train the mind}.}
        \label{fig:img1_grounded}
    \end{subfigure}
    
    \vspace{0.5em}
    \begin{subfigure}{\textwidth}
    \centering
        \includegraphics[width=0.5\linewidth]{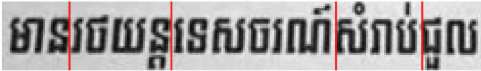}
        \caption{English translation: \emph{Tour vehicles for renting}.}
        \label{fig:img2_grounded}
    \end{subfigure}
    \vspace{0.5em}
    \begin{subfigure}{\textwidth}
    \centering
        \includegraphics[width=0.5\linewidth]{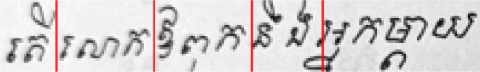}
        \caption{English translation: \emph{How about father and mother}.}
        \label{fig:img3_grounded}
    \end{subfigure}
    \vspace{0.5em}
    \caption{The projected word boundary locations on the input images. Red bar: boundary location.}
    \label{fig:grounded_evaluation}
\end{figure}

Figure~\ref{fig:grounded_evaluation} shows the projections of the word boundary locations on the input images. Except for highly curved text (e.g., Figure~\ref{fig:img3_grounded}), the projected word boundary locations are reasonably accurate. It should be noted that the visual encoder downsamples the features by a factor of four in the width direction, which makes it inevitable that the predicted locations may be slightly off.

In terms of the alternative design for the word boundary projector module, Tables~\ref{tab:reg_accuracy_gating} and~\ref{tab:reg_accuracy_gating_word} show that the default design of the word boundary projector module achieves better character recognition and word segmentation accuracies across the evaluation datasets. Thus, fusing the word boundary instruction via feature-wise linear modulation is more effective than simple gated feature fusion.

\begin{table*}[t]
\centering
\caption{Character error rates (CER) on the evaluation datasets of different modalities (D,S,\&H). * : model-specific tokenizer. Char.: character. KCC: Khmer character cluster. \textbf{Bold}: best. \emph{Italic}: second best.}
\label{tab:reg_accuracy_gating}
\begin{tabular}{lcccccccc}
\toprule
  & KHOB(D) & KhmerST(S) & KH(D,H) & GKST(S) & KHT(H) \\
\midrule
\textbf{KTRWS} (Gating; $b=0$) &  1.75 & 2.98 & 6.23 & 4.92 & 9.84\\
\textbf{KTRWS} (Gating; $b=1$)  &  1.82 & 3.27 & 7.62 & 5.17 & 10.46\\
\midrule
\textbf{KTRWS} (FiLM; $b=0$)  &  1.75 & 2.77 & 5.78 & 4.51 & 8.91\\
\textbf{KTRWS} (FiLM; $b=1$) &  1.79 & 3.04 & 6.06 & 5.01 & 9.64\\


\bottomrule
\end{tabular}
\end{table*}

\begin{table*}[t]
\centering
\caption{F1 scores on the evaluation datasets of different modalities (D,S,\&H). * : model-specific tokenizer. Char.: character. KCC: Khmer character cluster. \textbf{Bold}: best. \emph{Italic}: second best.}
\label{tab:reg_accuracy_gating_word}
\begin{tabular}{lcccccccc}
\toprule
 & KHOB(D) & KhmerST(S) & KH(D,H) & GKST(S) & KHT(H) \\

\midrule
\textbf{KTRWS} (Gating; $b=0$) &  98.48 & 97.73 & 94.06 & 96.23  & 93.01\\
\textbf{KTRWS} (Gating; $b=1$)  &  97.50 &  94.69 & 87.97 & 93.01  & 87.87\\
\midrule

\textbf{KTRWS} (FiLM; $b=0$) &  98.49 & 97.90 & 95.84 & 96.41  & 93.40\\
\textbf{KTRWS} (FiLM ; $b=1$)  &  97.68 &  95.08 & 91.09 & 93.35  & 89.24\\


\bottomrule
\end{tabular}
\end{table*}
\section{Limitations \& Future Work}

We identify the following limitations and future directions associated with this study:

\begin{enumerate}
\item The training relies on the new synthetic dataset with word boundary annotations. Although both recognition and segmentation performance are high for document images, there is room for improvement for other modalities (i.e., scene and handwritten). Thus, future work will focus on constructing a diverse real dataset with word boundary annotations.
\item The word boundary is implicitly handled by minimizing the $\mathcal{L}_{\text{CTC}}$ loss. Thus, future work will incorporate explicit word boundary supervision.
\end{enumerate}

\section{Conclusion}
We propose a novel joint Khmer text recognition and word segmentation (KTRWS) framework within a single model. With a CTC decoder for fast, parallel decoding, the KTRWS model can be instructed to recognize Khmer text with ($b=1$) and without ($b=0$) word segmentation. Experimental results on benchmark datasets of different document modalities (document, scene, and handwritten images) show that the proposed model not only recognizes characters in document images but also locates word boundaries, removing the need for an extra word segmentation step in a conventional sequential pipeline.


%
%
\bibliographystyle{splncs04}
\bibliography{main}
\end{document}